\documentclass[runningheads]{llncs}
\usepackage[T1]{fontenc}
\usepackage{graphicx}
\usepackage{algorithm}
\usepackage{algpseudocode}
\usepackage{amsmath,amsfonts,amssymb,bm}
\usepackage{booktabs}
\usepackage{subcaption}
\usepackage{xcolor}
\usepackage[numbers]{natbib}
\usepackage{hyperref}
\usepackage{multirow}
\begin{document}

\title{Adaptive Post-Processing Drives Instance-Level Detection in Stroke Lesion Segmentation}
\titlerunning{Adaptive Post-Processing Drives Instance-Level Detection}

% \begin{comment} %% Removed for anonymized MICCAI submission

\author{Qinghui Liu\inst{1}\thanks{Corresponding author} \and 
Jon Andr\'{e} Ottesen\inst{1} \and 
Atle Bj\o{}rnerud\inst{1} \and 
Kyrre Eeg Emblem\inst{1}}
\index{Liu, Qinghui}
\index{Ottesen, Jon Andr\'{e}}
\index{Bj\o{}rnerud, Atle}
\index{Emblem, Kyrre Eeg}

\institute{
Oslo University Hospital, Norway \\
\email{qiliu@ous-hf.no}
}
\authorrunning{Q. Liu et al.}
% \end{comment}

% \author{Anonymized Authors}  %% Added for anonymized MICCAI submission
% \authorrunning{Anonymized Author et al.}
% \institute{Anonymized Affiliations \\
%     \email{email@anonymized.com}}

\maketitle

\begin{abstract}
Instance-level lesion detection has been an increasingly larger focal point in medical image segmentation besides the more standard voxel-level overlap. Still, most pipelines are trained and post-processed for voxel overlap alone. In particular, the mismatch is most pronounced for small lesions, where a near-miss prediction—substantial overlap that falls just short of the instance-matching threshold—scores the same as a complete miss. In our ISLES'26 submission, we found that closing this gap mattered far more in post-processing than in architecture design. Our Volume-Conditioned Adaptive Post-Processing (VCAP) scheme adjusts component-size thresholds to each case's predicted lesion burden, improving Lesion-F1 by 0.032 (unbiased cross-fold estimate), approximately \(\sim\)6$\times$ larger than any architectural change we tested. A resolution-aware attention architecture (Viola2Plus), designed for small-lesion segmentation, shows why the distinction matters: it left small-lesion Dice unchanged but raised small-lesion detection rate by 3.7\%, a real effect voxel-overlap metrics alone would have missed. Under 5-fold cross-validation on the 1,453-case training set, our post-processed two-architecture ensemble achieves Dice 0.651 and Lesion-F1 0.614, versus 0.644 and 0.573 for the unprocessed single-model baseline.
\end{abstract}


\section{Introduction}\label{sec:intro}

Segmentation models are almost always trained to maximize voxel overlap between expert-based ground-truth annotations and model predictions, most commonly via Dice and cross-entropy losses. Computed over large volumes, these losses are dominated by large lesions -- a single 300\,ml lesion contributes far more gradient signal than a 0.05\,ml lacunar lesion. This training-time bias becomes a problem once a model is judged not just on how much it overlaps the ground truth, but on instance-level criteria such as lesion counts and total burden: a loss that barely notices small lesions produces a model that is correspondingly bad at finding them. The ISLES'26 challenge~\cite{isles26github,liew2022atlas} scores submissions across five metrics: Dice, instance-matched Lesion-F1 (IoU $\geq 0.25$), Absolute Lesion-count Difference (ALD), Absolute Volume Difference (AVD), and PR-AUC. 

\begin{figure}[!tbp]
\centering
\includegraphics[width=\textwidth]{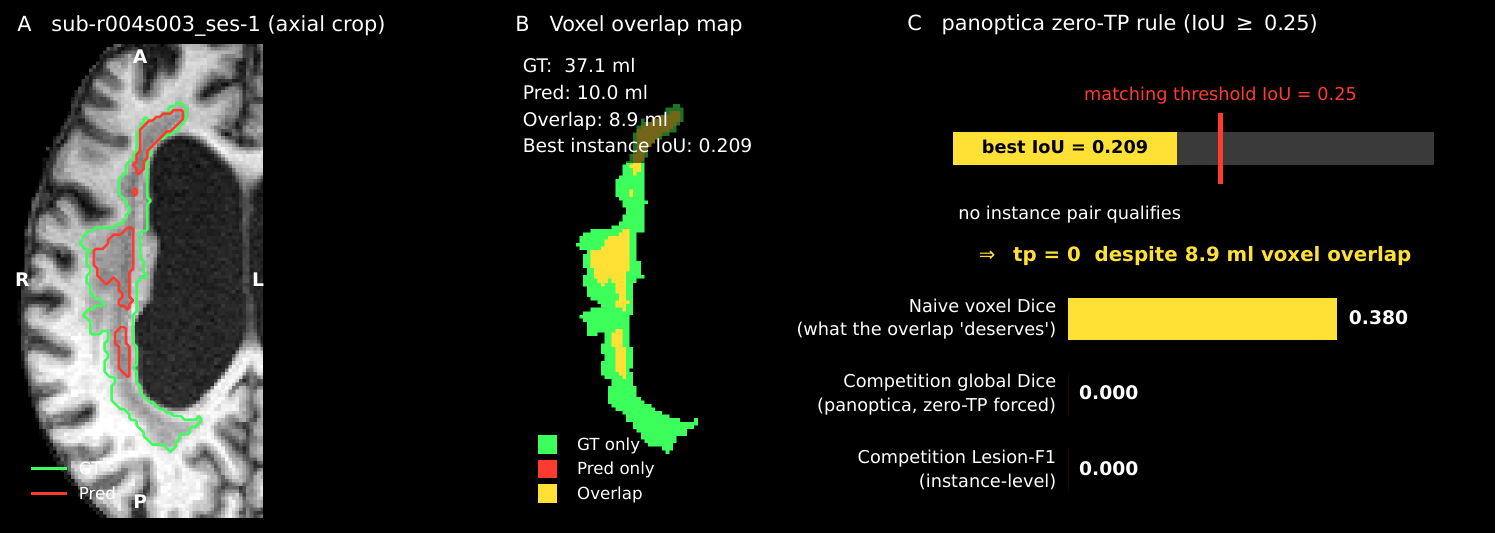}
\caption{\textbf{The instance-matching gate.} \emph{(A)} Axial crop: ground-truth (green) and predicted (red) contours overlap substantially along a thin, ventricle-hugging lesion. \emph{(B)} GT is 37.1\,ml, prediction is 10.0\,ml, overlap is 8.9\,ml -- yet the best instance-level IoU is only 0.209. \emph{(C)} Because 0.209 falls short of the instance-level matching threshold (IoU $\geq$ 0.25), no instance pair qualifies and tp $=$ 0: the naive voxel Dice this overlap would ``deserve'' is 0.380, but the competition-scored Dice and Lesion-F1 are both forced to 0.}
\label{fig:gate}
\end{figure}

Confirmed at the evaluation-library level~\cite{isles26github}, \emph{Dice and Lesion-F1 share the same instance-matching gate}: if no predicted component reaches IoU $\geq 0.25$ with any ground-truth lesion, Dice is forced to zero no matter how many voxels overlap (Figure~\ref{fig:gate}). Training on voxel-level Dice does nothing about this edge case on its own.

To address the challenges described above, our paper focuses on reducing the gap between the predicted lesion-burden and the scored lesion burden. Post-processing calibrated to predicted lesion burden is the most effective approach we found for closing the aforementioned problem, and the resulting configuration transfers across architectures without retraining. A stratified failure analysis also shows that an attention architecture can raise small-lesion \emph{detection} without changing voxel \emph{overlap} -- a reminder that voxel-level and instance-level gains are not interchangeable evidence of a method's performance.

\section{Methods}\label{sec:methods}

\subsection{Dataset and Preprocessing}\label{sec:dataset}
\begin{figure}[!tbp]
\centering
\includegraphics[width=\textwidth]{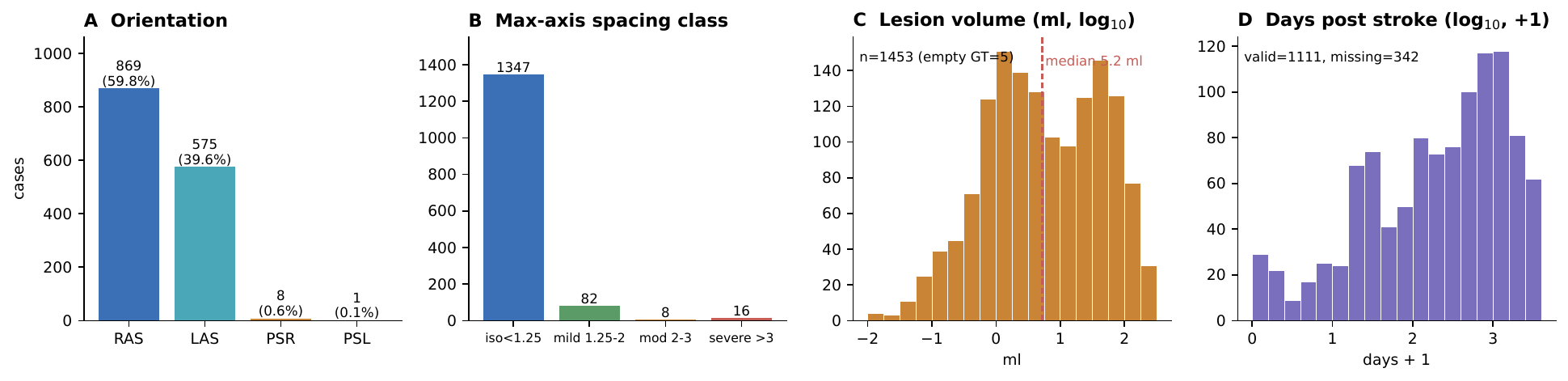}
\caption{\textbf{Dataset heterogeneity} ($n=1453$). \emph{(A)} Acquisition
orientation: predominantly RAS (59.8\%) and LAS (39.6\%). \emph{(B)}
Anisotropy: most scans are near-isotropic ($<$1.25\,mm), with 7.3\%
moderately to severely anisotropic. \emph{(C)} Lesion volume is heavy-tailed
(median 5.2\,ml), including 5 empty ground-truth cases. \emph{(D)} Time
since stroke onset, missing for 342 cases.}
\label{fig:landscape}
\end{figure}

The ISLES'26 training set comprises 1,453 multi-center T1-weighted MRI cases
from 72 acquisition cohorts (71 sites plus the SOOP cohort).
Figure~\ref{fig:landscape} summarizes the cohort's heterogeneity: acquisition
orientation, voxel-spacing anisotropy, lesion volume, and metadata
completeness all vary substantially -- time since stroke onset, for
instance, is missing for 23.5\% of cases. Intensity scale also varies by up
to three orders of magnitude across centers, which is why we normalize per
case rather than per cohort. All models use images resampled to isotropic
1\,mm spacing.

\subsection{Network Architecture}\label{sec:architecture}
We compare two architecture families under identical plans geometry, data,
and fold partitions (128$^3$ patch, batch size 2, deep supervision, 1\,mm
isotropic), so architecture is the only varied factor:
\begin{enumerate}
    \item \textbf{Baseline:} a stock nnU-Net
    \texttt{PlainConvUNet}~\cite{isensee2021nnunet,ronneberger2015unet}.
    \item \textbf{Viola2Plus:} deep decoder stages (320/256/128 channels)
    use tri-axial global attention over pooled axial statistics, in the
    spirit of Viola-UNet~\cite{liu2023ICH,Liu2026}. Shallow, high-resolution
    stages (64/32 channels) apply a local spatial gate in the spirit of the
    attention gates introduced for medical segmentation by Oktay et
    al.~\cite{oktay2018attentionunet}, modulated by upsampled deep features.
\end{enumerate}

All attention multipliers are initialized to the identity, so training
starts from the plain baseline and only learns deviations where the data
supports them. The local gate is boost-only, its multiplier bounded to
$[1,2]$, so it can amplify a region but never suppress one
(Section~\ref{sec:results-arch}). Both models use the standard nnU-Net
compound loss (Dice plus cross-entropy). %~\cite{milletari2016vnet} 

\subsection{Volume-Conditioned Adaptive Post-Processing (VCAP)}\label{sec:postproc}

A single global threshold and fixed minimum-component-size filter cannot
handle lesion burdens spanning several orders of magnitude: a permissive
filter (0.02\,ml) preserves lacunar infarcts but retains false positives
around large lesions, while a strict filter (0.1\,ml) cleans up large
infarcts yet discards small true positives. VCAP resolves this by making
the component-size threshold a function of the case's own predicted burden. 

% (Figure~\ref{fig:vcap_schematic})
% \begin{figure}[!tbp]
% \centering
% \includegraphics[width=\textwidth]{figures/fig_vcap_schematic_v3.pdf}
% \caption{\textbf{VCAP pipeline.} The ensemble soft map is binarized at a lowered threshold (0.35) to boost recall. The predicted lesion burden $V$~(ml) then drives a three-way router: empty rescue for near-zero burden ($V \leq 0.02$~ml), permissive dust removal ($\tau = 0.02$~ml) for moderate burden, and aggressive speck filtering ($\tau = 0.10$~ml) for high burden. All components are evaluated at 26-connectivity.}
% \label{fig:vcap_schematic}
% \end{figure}

Specifically, cases with predicted burden below 35\,ml use a permissive 0.02\,ml filter to retain small lesions, while cases above 35\,ml switch
to a stricter 0.1\,ml filter to suppress noisy fragments around large,
heterogeneous infarcts. A near-empty rule zeroes out any prediction whose
total retained volume falls below 0.02\,ml, targeting the empty-ground-truth edge case under PR-AUC. We deliberately lowered the binarization threshold from 0.5 to 0.35 to push under-segmented lesions past the IoU $\geq 0.25$
matching gate. All thresholds were jointly optimized via grid search on
pooled out-of-fold predictions, balancing all five metrics. Connected
components are computed with \texttt{cc3d}~\cite{silversmith2021cc3d} at
26-connectivity. The complete rule is stated compactly in Algorithm~\ref{alg:vcap}.

\begin{algorithm}[!tbp]
\caption{Volume-Conditioned Adaptive Post-Processing (VCAP)}
\label{alg:vcap}
\begin{algorithmic}[1]
\Require soft map $p \in [0,1]^{\Omega}$; voxel volume $v$ (ml);
  binarization threshold $\theta = 0.35$; burden tiers
  $(V_1, V_2) = (3.5, 35)$\,ml; per-tier minimum component volumes
  $(\tau_1, \tau_2, \tau_3) = (0.02, 0.02, 0.1)$\,ml; empty rule
  $\tau_{\mathrm{empty}} = 0.02$\,ml
\Ensure binary mask $b$; calibrated soft map $\tilde{p}$
\State $b \gets \mathbb{1}[\,p \geq \theta\,]$
  \Comment{recall-first binarization (lowered from 0.5)}
\State $V \gets |b| \cdot v$
  \Comment{total predicted burden for this case}
\If{$V < V_1$} $\tau \gets \tau_1$
\ElsIf{$V \leq V_2$} $\tau \gets \tau_2$
\Else{} $\tau \gets \tau_3$
\EndIf
\State $\{C_1, \dots, C_K\} \gets \textsc{ConnectedComponents}(b,\;
  26\text{-connectivity})$
\State $b \gets \bigcup_{k:\, |C_k|\cdot v \,>\, \tau} C_k$
  \Comment{volume-conditioned component filter}
\If{$|b| \cdot v \leq \tau_{\mathrm{empty}}$}
  \State $b \gets \mathbf{0}$;\; $\tilde{p} \gets \mathbf{0}$
  \Comment{empty rescue (empty-GT PR-AUC edge case)}
\Else
  \State $\tilde{p} \gets p$
\EndIf
\State \Return $b,\; \tilde{p}$
\end{algorithmic}
\end{algorithm}

\section{Results}\label{sec:results}

Official test-set labels are not available before submission, so all
reported metrics come from 5-fold cross-validation on the training set,
pooled across folds. The instance-matching gate (IoU $\geq 0.25$) is
reproduced using the organizers' \texttt{panoptica}
package~\cite{panoptica,isles26}. Table~\ref{tab:5fold} reports cross-validation performance before
post-processing or ensembling. Adoption of Viola2Plus followed a rule set
before we saw any results: adopt only if fold-0 Dice improved by at least
0.005 and Lesion-F1 by at least 0.01. Fold 0 cleared both bars ($+0.0105$
Dice, $+0.0212$ F1), and the remaining folds confirm the direction -- pooled
over all five, Viola2Plus improves every metric (Dice $+0.0045$, F1
$+0.0057$, PR-AUC $+0.0048$, AVD $-0.23$\,ml, ALD $-0.02$). 

\begin{table}[!tbp]
\centering
\caption{\textbf{5-fold cross-validation results, before post-processing.}
Baseline (Base) = stock nnU-Net PlainConvUNet; Viola2Plus (V2+) = dual-stage attention; identical data, folds, and plans geometry.}
\label{tab:5fold}
\resizebox{\textwidth}{!}{%
\begin{tabular}{l cc cc cc cc cc}
\toprule
\multirow{2}{*}{\textbf{Fold}} & 
\multicolumn{2}{c}{\textbf{Dice} $\uparrow$} & 
\multicolumn{2}{c}{\textbf{AVD (ml)} $\downarrow$} & 
\multicolumn{2}{c}{\textbf{ALD} $\downarrow$} & 
\multicolumn{2}{c}{\textbf{Lesion-F1} $\uparrow$} & 
\multicolumn{2}{c}{\textbf{PR-AUC} $\uparrow$} \\
\cmidrule(lr){2-3} \cmidrule(lr){4-5} \cmidrule(lr){6-7} \cmidrule(lr){8-9} \cmidrule(lr){10-11}
& Base & V2+ & Base & V2+ & Base & V2+ & Base & V2+ & Base & V2+ \\
\midrule
0 & 0.6569 & \textbf{0.6674} & \textbf{5.20} & 5.30 & \textbf{1.72} & 1.75 & 0.5811 & \textbf{0.6023} & 0.7546 & \textbf{0.7651} \\
1 & \textbf{0.6638} & 0.6636 & 6.87 & \textbf{6.36} & 2.39 & \textbf{2.28} & \textbf{0.5879} & 0.5873 & 0.7633 & \textbf{0.7652} \\
2 & \textbf{0.6231} & 0.6225 & 4.95 & \textbf{4.60} & \textbf{1.81} & \textbf{1.81} & \textbf{0.5766} & 0.5677 & \textbf{0.7210} & 0.7203 \\
3 & 0.6427 & \textbf{0.6516} & 5.91 & \textbf{5.86} & \textbf{1.73} & 1.77 & 0.5578 & \textbf{0.5652} & 0.7442 & \textbf{0.7556} \\
4 & 0.6343 & \textbf{0.6385} & 4.82 & \textbf{4.50} & 1.90 & \textbf{1.83} & 0.5608 & \textbf{0.5699} & 0.7364 & \textbf{0.7373} \\
\midrule
\textbf{Pooled} & 0.6442 & \textbf{0.6487} & 5.55 & \textbf{5.32} & 1.91 & \textbf{1.89} & 0.5728 & \textbf{0.5785} & 0.7439 & \textbf{0.7487} \\
\bottomrule
\end{tabular}%
}
\end{table}

\begin{figure}[!tbp]
    \centering
    \includegraphics[width=\textwidth]{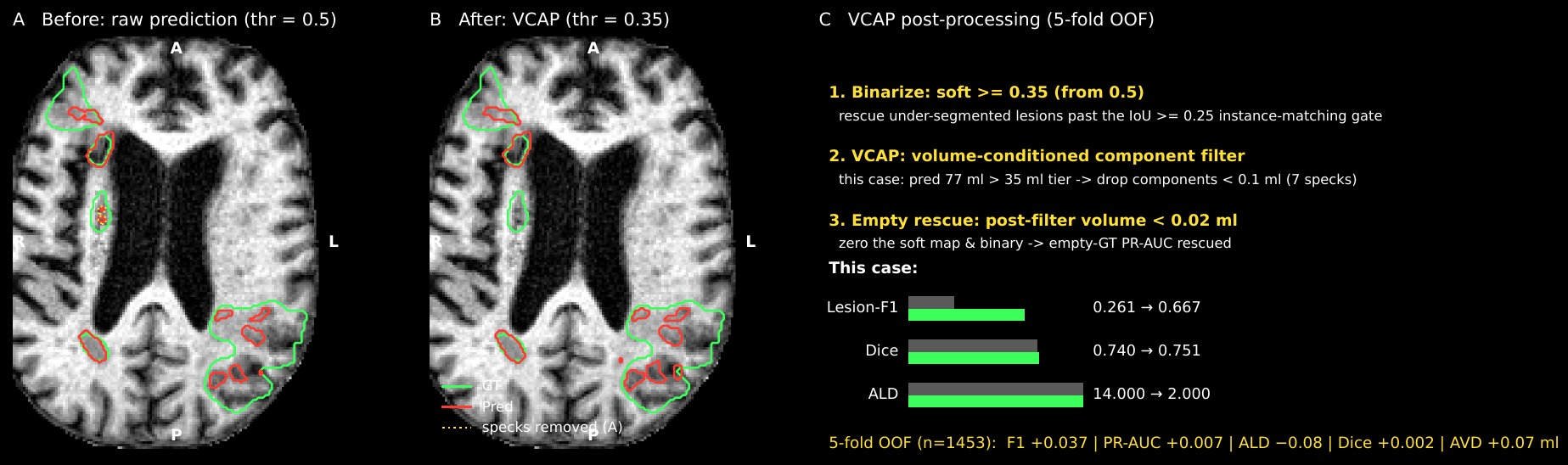}
    \caption{\textbf{Effect of VCAP on a thick-slice case.}
    \emph{Left:} raw prediction (threshold 0.5) with spurious fragments
    (yellow dotted). \emph{Center:} after VCAP, fragments removed
    (17 $\rightarrow$ 5 components). \emph{Right:} per-case metrics
    (Lesion-F1 0.261 $\rightarrow$ 0.667, ALD 14 $\rightarrow$ 2) alongside
    pooled 5-fold deltas.}
    \label{fig:vcap}
\end{figure}

\begin{table}[!tbp]
\centering
\small
\caption{\textbf{Post-processing ablation (pooled 5-fold, $n=1453$).}
VCAP $+$ = VCAP adds the near-empty zero-out rule. Ensemble = per-case average of baseline and Viola2Plus soft maps. The configuration was tuned on the baseline and applied unchanged to the others. }
\label{tab:postproc}
\begin{tabular}{lccccc}
\toprule
\textbf{Configuration} & \textbf{Dice} $\uparrow$ & \textbf{AVD (ml)} $\downarrow$ & \textbf{ALD} $\downarrow$ & \textbf{Lesion-F1} $\uparrow$ & \textbf{PR-AUC} $\uparrow$ \\
\midrule
Baseline & 0.6442 & 5.55 & 1.91 & 0.5728 & 0.7439 \\
Viola2Plus & 0.6487 & \textbf{5.32} & 1.89 & 0.5785 & 0.7487 \\
Ensemble     & 0.6488 & 5.37 & 1.85 & 0.5831 & 0.7556 \\
Baseline, VCAP+    & 0.6461 & 5.62 & 1.83 & 0.6095 & 0.7512 \\
Viola2Plus, VCAP+  & 0.6494 & 5.40 & \textbf{1.81} & 0.6122 & 0.7534 \\
Ensemble, VCAP+    & \textbf{0.6509} & 5.52 & 1.82 & \textbf{0.6143} & \textbf{0.7599} \\
\bottomrule
\end{tabular}
\end{table}

\subsection{VCAP Yields the Greatest Instance-Level Gains}\label{sec:results-postproc}

Figure~\ref{fig:vcap} shows this on one case: VCAP removes spurious
components, taking Lesion-F1 from 0.261 to 0.667 and ALD from 14 to 2.
Re-tuning the grid search on the Viola2Plus and ensemble outputs shifts the
optimal binarization threshold from 0.35 to 0.5, but gains at most 0.002 on
any metric over the original configuration -- a wide, flat optimum -- so we
kept one configuration for all three families (Table~\ref{tab:postproc}).

VCAP improves cohort-level Lesion-F1 by 0.037 over the unfiltered baseline
(0.5728 $\to$ 0.6095) under full-data selection; under an unbiased
leave-one-fold-out nested-selection protocol, the estimate is 0.032, stable
across folds -- the number we treat as the honest effect size. The gain
holds across all three model families, though it shrinks slightly as the
underlying model improves: VCAP adds $+0.0367$ F1 to the baseline,
$+0.0337$ to Viola2Plus, and $+0.0312$ to the ensemble, consistent with
post-processing and model quality partially overlapping in what they fix.
This is not free: AVD rises by 0.07\,ml relative to the unfiltered baseline,
the cost of lowering the binarization threshold to rescue under-segmented
lesions past the matching gate.

\begin{figure}[!tbp]
\centering
\includegraphics[width=\textwidth]{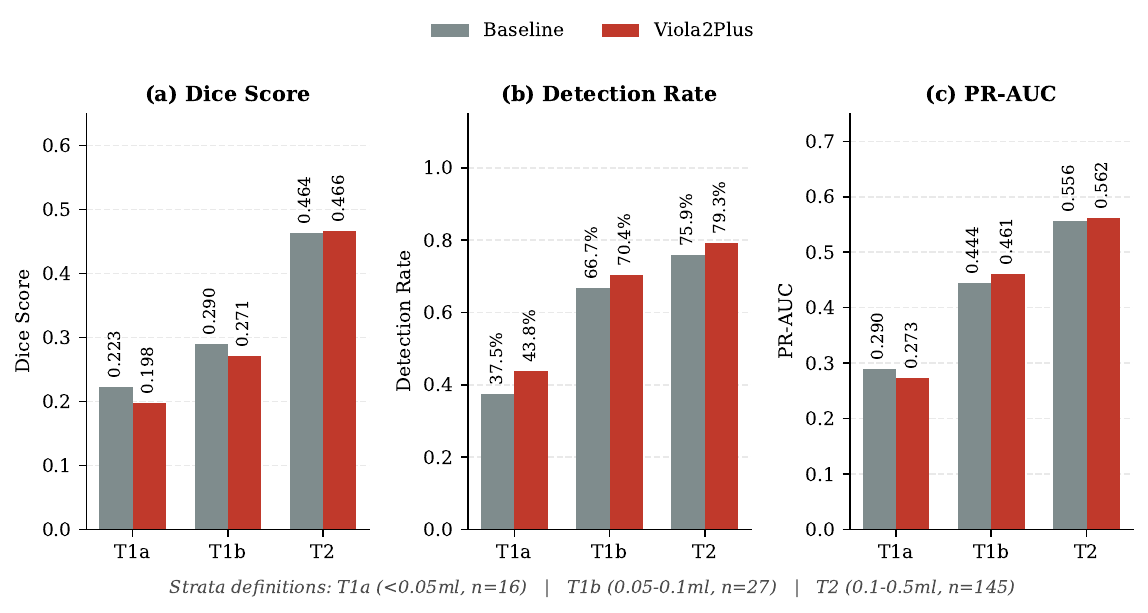}
\caption{\textbf{Performance on small-lesion strata (pooled 5-fold, no post-processing).} Lesions are categorized by ground-truth volume: T1a ($<$0.05\,ml), T1b (0.05--0.1\,ml), and T2 (0.1--0.5\,ml). While Viola2Plus demonstrates a clear advantage in instance-level retrieval---consistently boosting \textbf{(b)} Detection Rate across all tiers and improving \textbf{(c)} PR-AUC for lesions $>0.05$\,ml---it yields no corresponding benefit in voxel-level overlap, as evidenced by the stagnant or slightly degraded \textbf{(a)} Dice scores. This highlights the architecture's capacity to amplify weak signals for detection without refining spatial boundaries.}
\label{fig:main}
\end{figure}

\subsection{Architectural Nuance: Enhancing Detection, Not Overlap}\label{sec:results-arch}

Viola2Plus was designed to specifically improve small-lesion segmentation, but its empirical behavior reveals a more nuanced dynamic (Figure~\ref{fig:main}). Across the 188 small-lesion cases ($<$0.5\,ml), the overall detection rate rises by 3.7 points (0.713 to 0.750)---about three times the architecture's cohort-wide detection improvement. 

When stratified by volume, Viola2Plus consistently increases instance-level detection across all tiers, with the most pronounced gain in the extremely challenging T1a tier ($<$0.05\,ml, $+6.3\%$ points). Notably, the precision-recall AUC (PR-AUC) also improves for T1b and T2 lesions, indicating that this heightened sensitivity does not come at the severe cost of precision. However, this enhanced retrieval ability does not translate to better voxel-level overlap: small-lesion Dice scores remain flat or slightly decrease (overall $-0.003$, $p=0.25$), and AVD gets marginally worse ($+0.19$\,ml). 

This behavioral pattern is highly consistent with the local gate's boost-only design. A mechanism that can amplify a candidate region but never suppress one effectively pushes borderline small lesions past the detection threshold---improving both detection rate and PR-AUC---but it lacks the inherent capability to tighten their spatial boundaries, resulting in stagnant Dice scores and the observed AVD increase.

\section{Discussion and Conclusion}\label{sec:discussion}
Our volume-conditioned adaptive post-processing (VCAP) approach produced an instance-level gain roughly 6 times larger than any architecture change we tried (0.032 vs.\ 0.006 Lesion-F1). The two levers are largely independent: gains from ensembling, post-processing, and architecture stack rather than compete. A single global threshold implicitly assumes a scale-invariant trade-off between keeping true positives and rejecting false ones, which breaks down whenever lesion size spans several orders of magnitude -- common in stroke imaging and probably elsewhere. Once calibrated, VCAP transfers across model families with negligible loss.

The small-lesion result makes the same point from another angle: an architecture built to improve overlap instead shifted detection upward while leaving overlap flat -- something a Dice-only evaluation would have read as no effect at all. Wherever Dice and an instance-matching metric share a detection gate, a change can do something real that overlap metrics alone cannot see, so architecture claims under instance-aware metrics are worth checking against detection rate and volume bias directly, not inferred from Dice.

Two limitations are relevant for these conclusions. The empty-ground-truth subgroup is small (5 of 1,453 cases) and cannot support a strong claim either way. VCAP's gain is also not free: AVD rises by 0.07\,ml, the cost of lowering the binarization threshold to rescue borderline lesions. Our 5-fold internal estimate, finally, is not an official test-set result.

% The broader lesson: check what your evaluation metric rewards before you
% redesign your network. Overlap gains do not guarantee instance-level
% gains.

\begin{credits}
\subsubsection{\ackname}
The authors acknowledge support from the Helse Sør-Øst regional health authority of Norway (Grant 2021031).

\subsubsection{\discintname}
The authors have no competing interests to declare that are relevant to the content of this article.
\end{credits}

\bibliography{references}

\end{document}